\pdfoutput=1

\documentclass[11pt]{article}

\usepackage[preprint]{acl}

\usepackage{times}
\usepackage{latexsym}
\usepackage{float}
\usepackage[T1]{fontenc}

\usepackage[utf8]{inputenc}

\usepackage{microtype}

\usepackage{inconsolata}

\usepackage{graphicx}

\title{Probing Language Models' Gesture Understanding \\ for Enhanced Human-AI Interaction}

\author{Philipp Wicke \\
  Center for Information and Language Processing, LMU Munich \\
  Munich Center for Machine Learning (MCML) \\
  \texttt{pwicke@cis.lmu.de} }

\begin{document}
\maketitle
\begin{abstract}
The rise of Large Language Models (LLMs) has affected various disciplines that got beyond mere text generation. Going beyond their textual nature, this project proposal aims to investigate the interaction between LLMs and non-verbal communication, specifically focusing on gestures. The proposal sets out a plan to examine the proficiency of LLMs in deciphering both explicit and implicit non-verbal cues within textual prompts and their ability to associate these gestures with various contextual factors. The research proposes to test established psycholinguistic study designs to construct a comprehensive dataset that pairs textual prompts with detailed gesture descriptions, encompassing diverse regional variations, and semantic labels. To assess LLMs' comprehension of gestures, experiments are planned, evaluating their ability to simulate human behaviour in order to replicate psycholinguistic experiments. These experiments consider cultural dimensions and measure the agreement between LLM-identified gestures and the dataset, shedding light on the models' contextual interpretation of non-verbal cues (e.g. gestures).
\end{abstract}

\section{Introduction}

The successful launch of OpenAI's ChatGPT in November 2022, with one million users within five days \cite{hu2023chatgpt}, sparked considerable interest in conversational AI. Built on the Generative Pretrained Transformer (GPT) architecture \cite{radford2019language,brown2020language}, ChatGPT employs attention mechanisms for text generation in a dialogue-format. Trained extensively, underlying Large Language Models (LLMs) demonstrate rudimentary forms of creativity and comprehension, prompting speculation about ``sparks of general artificial intelligence'' \cite{bubeck2023sparks}. This raises questions about whether LLMs simulate language and exhibit cognitive elements akin to machine or human cognition. Distinguishing human cognition from machines connects to the philosophical concept of Multiple Realizability \cite{putnam1967psychological}, suggesting that different physical systems can produce similar cognitive processes. 

Embodied cognition, emphasizing the role of an organism's body and sensory experiences \cite{varela2017embodied}, adds to this perspective. Exploring how embodied experiences influence LLMs' cognitive attributes becomes crucial as scientific and non-scientific domains increasingly leverage LLM capabilities \cite{biswas2023chatgpt,van2023chatgpt,wu2023bloomberggpt}.
Simultaneously, addressing bias within language models is essential due to concerns about perpetuating societal biases \cite{tamkin2021understanding}. Given the absence of a physical body in LLMs, questions arise about the presence or absence of fundamental principles governing human-AI interactions in their representations. These inquiries necessitate multifaceted exploration. 

\begin{figure}
    \centering
    \includegraphics[width=.5\textwidth]{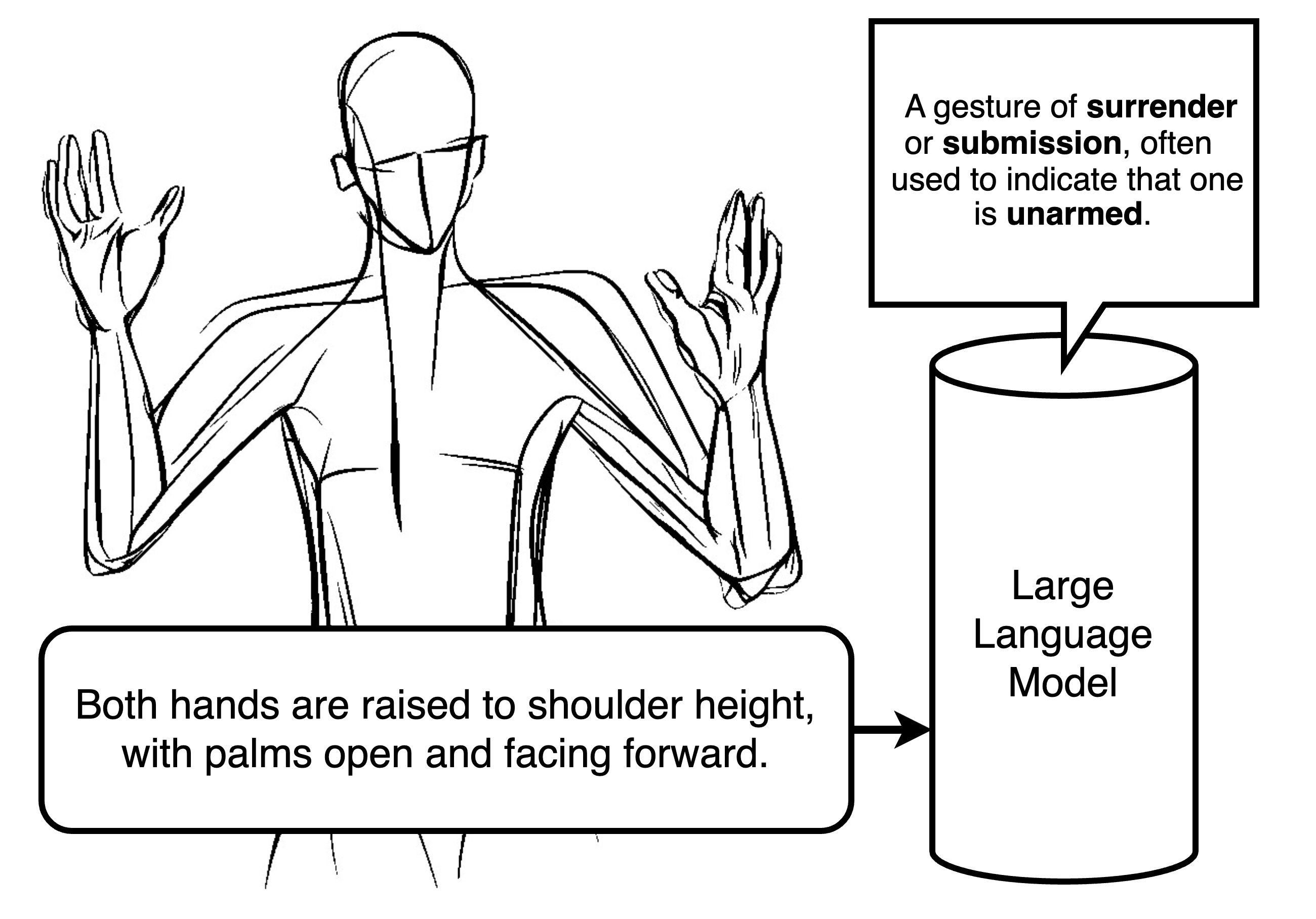}
    \caption{Probing a Large Language Model (LLM) through the input of gesture descriptions can serve as a valuable means to evaluate its understanding of gestures, contributing to the refinement of human-AI interaction.}
    \label{fig:enter-label}
\end{figure}

The proposed research complements ongoing efforts exploring these questions. Related research investigates embodying LLMs in robots for enhanced human-robot interactions \cite{wicke2023towards}. Additionally, an examination of LLMs' interpretation of figurative language, especially metaphors, reveals perceived embodiment significantly aids their interpretative capacity \cite{wicke2023lms}. Adding a third perspective, the proposed project focuses on the spatial dynamics of human non-verbal communication, namely gestures. Gestures, crucial for structuring communication \cite{mcneill1992hand}, bridge the gap between linguistic concepts and bodily expressions \cite{mittelberg2006metaphor}. Currently, no interdisciplinary exploration exists into the role of gestures in the LLM context. Important questions remain: 
\begin{quote}
    How do LLMs conceptualize gestures, and can they accurately interpret gestural cues translated into text?
\end{quote}
This research aims to fill this gap, offering insights into the comprehension, interpretation, and potential utilization of non-verbal communication cues by LLMs.

\section{Related Works}

\paragraph{Computational Representations of Gestures} Historically, gesture representation has been a fundamental aspect of understanding human communication and interaction with computers. Early works, such as those by \citet{mcneill1992hand} laid the foundation for analysing the linguistic and semiotic aspects of gestures. More recently, scholars such as \citet{cienki2008metaphor} are at the forefront of human gesture studies. \citet{cienki2008metaphor} emphasise the embodied and dynamic nature of gestures, viewing them as integral components of language that convey meaning through their interaction with speech and the surrounding context. This perspective aligns with the idea that gestures are not mere embellishments but constitute an 
essential part of the communication process. Regarding mental representations, investigates the endurance of iconic origins in emblematic gestures. \citeauthor{bergen2019gestures}'s studies suggest that emblematic gestures, despite their historical associations, undergo cognitive changes over time, highlighting the dynamic, cultural nature of the relationship between gestures and mental representations \cite{bergen2019gestures}. Computational representations on gestures can be found in work by  \citet{wicke2020show}. The work provides a taxonomy of schematic movements and gestures, which can be used to implement a variety of creative performance types with robots with an emphasis on an apt use of spatial movement.

\paragraph{LLMs and Robotic Bodies} Large Language Models are opaque statistical systems, which do not allow us, as opposed to word embeddings, to simply look up how certain concepts are defined. Moreover, their high-dimensional latent space may define ``gestures'' as multimodal as we humans conceptualise them. Hence, we can probe LLMs with datasets of tests, which require a certain concept to be present. For example, in research by \citet{wicke2023lms} tests LLMs ability to interpret figurative language with the FigQA dataset \cite{liu2022testing}. 

Moreover, to test the effect of embodiment on LLMs’ ability to understand metaphors, which are assumed to be derived from bodily experiences \cite{lakoff2008women}, the study correlates the performance of metaphor interpretation with the perceived embodiment of the action words \cite{sidhu2014effects} in each metaphor. The results suggest that the degree of embodiment has a positive impact on LLMs ability to do the correct interpretation. Similarly, the proposed project follows a comparable approach by testing LLMs' aptitude to accurately comprehend and attribute gestures to text, shedding light on the underlying conceptualisation within these text-based models.

\begin{figure*}[ht]
    \centering
\includegraphics[width=\textwidth]{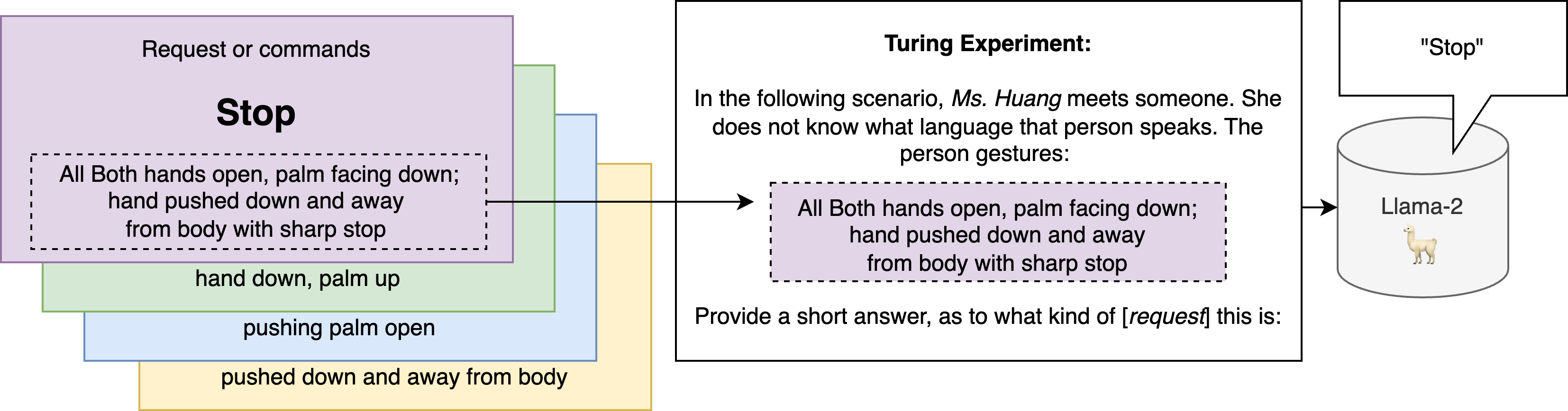}
    \caption{Suggested Turing Experiment (TE) based on the VLM list. The item from the VLM list  (e.g. \textbf{Stop}) is turned into an appropriate prompt for the TE, which is then fed to the language model (e.g. Llama-2) for evaluation.}
    \label{fig:texp}
\end{figure*}

\section{Methodology}
\paragraph{Research Questions} Motivated by the need for a systematic and comprehensive exploration, the research plan for this study is designed to shed light on the relationship between LLMs and non-verbal communication (e.g. gestures). This investigation has a bipartite structure: First, the most powerful types of models will be investigated, namely LLMs of different sizes (e.g. Llama-2, GPT-3). Even though these models are monomodal (i.e. textual), they provide the basis for multimodal models, which can be investigated in a second step, extending the project. Related works have already used multimodal models to create systems for gesture classification and generation \cite{ao2023gesturediffuclip,gao2023gesgpt}. It is important to limit the initial study to textual models, because current multimodal models are using smaller LLMs as a backbone structure, which inevitably guides any further multimodal conceptualisation. Hence, the following first set of pivotal questions can be derived:
\begin{itemize}
    \item \textbf{Accuracy in Interpretation:} How effectively can LLMs decipher explicit and implicit non-verbal cues (i.e. gestures) embedded within textual prompts?
    \item \textbf{Coherence in Cultural Contexts:} How adept are LLMs at associating these gestures with different contextual backdrops, revealing the depth of their comprehension?
\end{itemize}

\paragraph{Model selection} It is crucial to note that important LLMs (e.g. GPT-3) have been developed by corporations, exemplified by entities like OpenAI. However, an inherent limitation of these models is their unavailability with respect to model weights and the exact training procedures employed. In the pursuit of fostering transparency, reproducibility, and embracing the principles of open-source research, we choose to abstain from relying on these proprietary models. This approach not only facilitates transparency but also encourages the broader scientific community to engage with and build upon our work. The selection and comparison encompasses a diverse range of model parameter sizes, including (currently) popular open-source options such as Llama-2 \cite{touvron2023llama}, OPT \cite{zhang2022opt}, Alpaca \cite{zhang2023llama}, GPT-NeoX \cite{black2022gpt}, Bloom \cite{workshop2022bloom}, as well as others that may become available during the course of our research. For a potential multimodal extension, the selection of VLMs includes the open-source models InstructBlip \cite{dai2023instructblip} and OpenFlamnigo \cite{awadalla2023openflamingo}.

\paragraph{Dataset Selection} A fundamental issue of gestural data is that there cannot be a gold standard for the use of gestures, i.e. there is no one correct gesture accompanying a specific sentence. To begin with, gestures are not needed for communication, but can provide helpful additional semantic information. Yet, they show a range of profound socio-cultural differences \cite{matsumoto2013cultural}. Fortunately, a plethora of research in gesture studies provides appropriate 
psycholinguistic study designs that can aid the dataset construction. \citet{matsumoto2013cultural} present a Verbal Message List (VML), which includes 96 items ``deemed important for individuals interacting with people from a different culture for the first time to know in order to highlight emblematic differences'' \cite{matsumoto2013cultural}. These items include expressions such as \textit{Catastrophe}, \textit{Girlfriend}, \textit{I’m very strong} and their list is publicly available. The list comes categories (e.g. insult, request), the verbal message (e.g. \textit{Girlfriend}), region (e.g. \textit{global}, \textit{East Asia}) and a description of the gesture (e.g. \textit{Thumb of one hand out, other fingers curled; thumb
pointing in a desired direction}). This data has several benefits over other available datasets. Many recent text/gesture datasets rely on video annotations using the ELAN (EUDICO Linguistic Annotator) system \cite{zheng2022elan}. These datasets \cite{turchyn2018gesture,ienaga2022semi} rely on annotations of videos at certain temporal intervals and are mostly confined to the generation of new gestures without classification or explicit conceptualisation of gesture types. The VML will be used to curate a diverse dataset pairing textual prompts with detailed gesture descriptions, covering a wide spectrum of non-verbal cues, regional labels and semantic labels (e.g. request, insult etc.). Hence, the VML is chosen as the basis for the evaluation.

\paragraph{Turing Experiments} \citet{aher2023using} present the Turning Experiment (TE), which is a novel test used to assess the ability of LLMs to simulate various aspects of human behaviour, including replicating classic experiments in psycholinguistics. TEs involve simulating a representative group of human research participants with different cultural backgrounds within prompts given to the LLM.

These experiments aim to uncover how well LLMs can reproduce established research findings. These TEs enable to assess multiple different answers for gesture conceptualisation and can provide a cultural dimension. The performance of the LLMs can thus be assessed on the VML and the metric will be the amount of overlap between the LLMs identified gesture and the VML based on its description and context (see Figure \ref{fig:texp}). \\

In Figure \ref{fig:texp}, the VML provides 96 items (abstracted with coloured cards on the left) that are used to construct the individual TEs. Each experiment features a scenario description (white box in the center) with a varying gender (Mr./Mrs./Mx.) and cultural parameter. For example, Fig. 1 shows Ms. Huang as a female, East Asian participant. The response of the LLM is compared with the VLM label and a standard accuracy measure is conducted to assess the LLM performance on the task. Lastly, all LLMs will be compared and correlations between the cultural parameters will be assessed with respect to the conceptualisation of the gestures provided by the VLM semantics.

\section{Discussion}

The proposed research represents a pioneering exploration into the interaction between Large Language Models (LLMs), such as Llama-2 and GPT-3, and non-verbal communication cues, particularly gestures. By aiming to unravel how LLMs conceptualize and interpret gestures, the research delves into a novel and interdisciplinary domain at the intersection of artificial intelligence and human communication dynamics. The primary objective is to discern the proficiency of LLMs in comprehending and accurately representing non-verbal cues within textual prompts in order to lay the ground work for further studies intersecting the field of gesture studies and statistical language modelling.
\\

The planned investigation acknowledges the importance of cultural context in gesture comprehension. The findings should acknowledge that LLMs, when exposed to diverse datasets, may exhibit a commendable variability to associate gestures with different cultural backgrounds. This cultural sensitivity aligns with the broader aim of developing more inclusive and contextually aware conversational AI systems. Its insights may prove valuable for the future design on human-robot interaction that utilises generative language models.

\paragraph{Limitations}

The reliance on open-source models, while promoting transparency, may introduce limitations in terms of model complexity compared to proprietary counterparts like GPT-3. This raises questions about the generalizability of the findings to models with different architectures and scales. Moreover, the initial focus on monomodal (textual) models limits the exploration of multimodal capabilities, which could potentially enhance gesture comprehension. Future research may need to address this limitation to provide a more comprehensive understanding of non-verbal communication with artificial agents using generative models.

Lastly, as a proposal, the research has not been conducted, and the strengths and weaknesses outlined are based on anticipated outcomes. The actual performance of LLMs in gesture comprehension remains to be empirically tested and evaluated. 

\paragraph{Future Work}
Future research directions could include extending the study to multimodal models, exploring the integration of visual information to enhance gesture comprehension. Additionally, investigating the impact of gesture-based communication on user experience and engagement with conversational AI systems could provide valuable insights for the design and development of more effective and user-friendly interfaces. Moreover, all empirical evidence from the dataset needs to be verified in real-world robotic experiments.

\section{Conclusion}

The proposed research aims to establish a pioneering investigation between the domains of gesture studies, cognitive linguistics, and computational language modeling. This approach not only identifies prospective models but also delineates a pertinent dataset. The research design outlines the methodology for an empirical investigation, promising insights into the depth of LLMs' comprehension of gestures, with unclear outcomes. 

The exploration of various perspectives within the plan underscores the potential of this study to transcend its immediate context. The results of this interdisciplinary inquiry may serve as a catalyst for nuanced and multifaceted investigations. By interweaving insights from multiple disciplines, this research seeks to contribute substantively to the evolving discourse on human-AI interaction that rely on multimodal generative AI systems.

\section*{Acknowledgement}
\begin{itemize}

\item Figure 1 Image AI generated: Prompt = ``Simplistic abstract drawing of a figure ((gesturing)) on white background. clean lines.'', Model = stable-diffusion (sdxl), Guidance Scale = 7.5, Seed = 154221118039499.
\item Figure 1 Gesture Interpretation: Partial answer of Bing's Copilot.

\item Main Text: Formatting, grammar, syntax and writing assistance with OpenAI's ChatGPT.
\end{itemize}

\bibliography{custom}

\end{document}